\documentclass[conference]{IEEEtran}

\usepackage{cite}
\usepackage{amsmath,amssymb,amsfonts}
\usepackage{graphicx}
\usepackage{textcomp}
\usepackage{xcolor}
\usepackage{url}
\usepackage{booktabs}
\usepackage{multirow}
\usepackage{subcaption}
\usepackage{array}
\usepackage{tikz}
\usetikzlibrary{positioning, arrows.meta, shapes.geometric, fit, calc, backgrounds}

\definecolor{cInput}{HTML}{E8DEF8}     
\definecolor{cProc}{HTML}{E3F2FD}      
\definecolor{cStore}{HTML}{E0F2F1}     
\definecolor{cRouter}{HTML}{FFE8B5}    
\definecolor{cLLM}{HTML}{FCE4EC}       
\definecolor{cAnswer}{HTML}{C8E6C9}    
\definecolor{cBorder}{HTML}{1F2D3D}    
\definecolor{cEdge}{HTML}{37474F}      
\definecolor{cFrame}{HTML}{90A4AE}     

\begin{document}

\title{LogRouter: Adaptive Two-Level LLM Routing for\\ Log Question Answering in Big Data Systems}

\author{
\IEEEauthorblockN{Mert Coskuner}
\IEEEauthorblockA{
\textit{TUBITAK BILGEM} \\
Kocaeli, Turkey \\
mert.coskuner@tubitak.gov.tr}
\and
\IEEEauthorblockN{Merve Zeybel}
\IEEEauthorblockA{
\textit{TUBITAK BILGEM} \\
Kocaeli, Turkey \\
kaya.merve@tubitak.gov.tr}
\and
\IEEEauthorblockN{Melik Mert Dolan}
\IEEEauthorblockA{
\textit{TUBITAK BILGEM} \\
Kocaeli, Turkey \\
mert.dolan@tubitak.gov.tr}
}

\maketitle

\begin{abstract}
Production log analytics in self-hosted, resource-constrained
environments requires natural-language access to massive log streams without
the cost of routing every query through a large language model. We
present \textbf{LogRouter}, an end-to-end log question-answering system
deployed on TUBITAK BILGEM's national big data platform
that combines a PySpark-based Drain3 ingestion
pipeline, GPU-accelerated embeddings, and dual-index storage in Apache
Druid and PostgreSQL with \texttt{pgvector}. A
two-level cost-aware router dispatches each query along one of four
execution paths---direct response, Druid keyword search, template
lookup with SQL generation, and \texttt{pgvector} semantic
retrieval---while a Level-2 router selects a 14B-class or 32B-class
generator for the semantic path. A dedicated coder LLM handles
text-to-SQL.
We evaluate the system on four LogHub datasets (Linux, Apache, Windows,
Mac; 70 questions in total) under both an online full-pipeline
configuration and an offline configuration that isolates the
generator. The router reaches \textbf{88.4\,\%} mean accuracy across
datasets and \textbf{94.7\,\%} on Linux, while the full pipeline attains
a mean ROUGE-1 of 0.373, BERTScore of 0.879, RAGAS Faithfulness of
0.779, and an end-to-end latency of 18.6\,s. In an apples-to-apples
offline comparison, the routed system cuts mean latency by
\textbf{55\,\%} versus a Fixed-32B baseline (46.3\,s vs.\ 102.1\,s)
while preserving Answer Correctness within 5.8 points and exceeding a
Fixed-14B baseline on RAGAS Faithfulness across every dataset.
Cost-aware dispatching is therefore a practical mechanism for
production log QA: routing recovers most of the quality of an
always-32B configuration at less than half the latency, and the L1
keyword vocabulary makes that routing decision with high precision
without a learned classifier.
\end{abstract}


\section{Introduction}
Operational log analysis is a core software maintenance activity:
maintainers continually consult log streams to comprehend system
behaviour, diagnose regressions, and investigate incidents on live
deployments. Modern distributed systems emit these streams at a rate
that outpaces human inspection, and maintainers routinely need
natural-language access during evolution and incident response.
Retrieval-augmented generation
(RAG)~\cite{lewis2020rag} provides a compelling interface, yet
deploying large language models (LLMs) over operational log data in a
real enterprise environment surfaces three practical challenges that
the maintenance and evolution literature has not jointly addressed.
This work reports our experience deploying such a system within
TUBITAK BILGEM, a national research institute that operates a
production big data platform for public-sector analytics workloads.
This paper contributes to the use of \emph{Large Language Models for
software evolution and maintenance tasks}, providing an empirical
evaluation of an on-premises LLM-based log QA system in a live
production setting.

\textbf{Challenge 1: Scale and continuity.}
Existing log-RAG prototypes operate on static, pre-collected log
files~\cite{mudgal2023chatgptlogs, qi2023loggpt}. A production platform
cannot batch-process logs offline: ingestion must be continuous, parsing
must be distributed, and the indexes that feed retrieval must remain
fresh. In our deployment on TUBITAK BILGEM's national big data
platform, we meet this requirement with a
streaming pipeline spanning Grafana Loki, PySpark-based Drain3 template
extraction~\cite{he2017drain}, GPU-accelerated embedding via Ollama, and
dual-index storage in Apache Druid (structured/keyword) and PostgreSQL
with \texttt{pgvector} (semantic).

\textbf{Challenge 2: Heterogeneous query types and LLM cost.}
Production log workloads contain fundamentally different question types.
\emph{``How many \texttt{ERROR} events occurred in the last hour?''} is
a database aggregation; \emph{``Why did the ingestion pipeline stall on
Tuesday?''} requires multi-hop semantic reasoning over many log lines.
Early iterations of our system routed every query through a single large
generative model, wasting GPU resources on structured lookups and
returning numerically incorrect counts instead of exact database
results. The deployment insight is that \emph{query type, not query
content, determines the appropriate backend}: misrouting a simple
fact-seeking query to an LLM is not just inefficient, it is actively
harmful to answer quality.

\textbf{Challenge 3: Right-sizing the generator.}
Even within the semantically-routed subset, not all questions are
equal. Summarising a failure pattern across fifty log lines requires
a capable model; identifying the service that emitted a single line
does not. A static one-model policy either wastes capacity on simple
queries or degrades under concurrent workloads on a single GPU.

We address these challenges with \textbf{LogRouter}, a production log
QA system deployed on TUBITAK BILGEM's national big data platform.
Building on the
modular routing philosophy of LLMLogAnalyzer~\cite{cai2025llmloganalyzer},
LogRouter extends it to a distributed big data environment with three
practical extensions: (i) explicit LLM-bypass paths that skip generative
inference entirely for non-semantic queries; (ii) Level-2 dynamic
model-size selection that assigns the 32B generator only where complexity
warrants it; and (iii) fully self-hosted GPU inference with no
external API calls, a hard requirement in public-sector deployments
where log data may not leave the organisation.

The main contributions of this paper are:
\begin{itemize}
    \item An end-to-end log analysis pipeline that connects Grafana
    Loki ingestion, PySpark-based Drain3 template
    mining~\cite{he2017drain}, GPU-accelerated chunk embedding via
    Ollama, and dual-index storage in Apache Druid and \texttt{pgvector}.

    \item A two-level cost-aware query router: Level-1 dispatches
    across four execution paths (\textsc{General}, \textsc{Keyword},
    \textsc{SQL}, \textsc{Semantic}) using a regex-based keyword signal
    vocabulary of seven structural patterns (P0--P6) and a
    question-starter guard (P7); Level-2 selects the generator size
    (14B or 32B) for the semantic path; and a dedicated coder LLM serves
    the SQL path.

    \item An evaluation on four LogHub datasets (70 questions) under
    both online (Druid + \texttt{pgvector}) and offline (raw chunk
    injection) configurations, including a nine-condition ablation
    study covering router removal, hybrid retrieval, model-size
    fixation, and Drain removal.

    \item Deployment lessons: (a) routing recovers most of the
    faithfulness of an always-32B configuration at less than half the
    latency; (b) Drain template extraction contributes almost entirely
    to routing precision, not to generation quality; (c) LLM-based
    judges systematically penalise the verbatim style of bypass
    answers, a structural artefact rather than a quality regression.
\end{itemize}

\section{Related Work}

\subsection{Log Parsing}
Automated log parsing extracts structured templates from unstructured
log lines and is a prerequisite for downstream analytics.
Drain~\cite{he2017drain} clusters log lines online with a fixed-depth
parse tree, achieving high accuracy at low overhead; the production
fork Drain3 adds configurable masking and persistence. Benchmarks on
LogHub~\cite{zhu2023loghub} confirm Drain's strong accuracy and
throughput. All published Drain3 deployments operate on single-node,
file-based collections; to our knowledge no prior work runs Drain3 as a
PySpark stage over live log streams, the gap our ingestion architecture
addresses.

\subsection{Log Analysis and Anomaly Detection}
A substantial body of work addresses log-based anomaly detection.
DeepLog~\cite{du2017deeplog} models log-event sequences with an LSTM;
LogBERT~\cite{guo2021logbert} fine-tunes BERT on log templates;
LogRobust~\cite{zhang2019logrobust} adds semantic representations for
robustness to log evolution. All three share the same design goal of
classifying sequences as normal or anomalous; none support free-form
natural-language querying or the heterogeneous mix of keyword,
aggregation, and semantic queries that operators ask in practice. Our
system is complementary: rather than detecting anomalies, it provides
an interactive QA interface over historical log data.

\subsection{LLM-Based Log Analysis}
LogGPT~\cite{qi2023loggpt} applies ChatGPT to log anomaly detection
with strong interpretability but full reliance on commercial cloud APIs.
Mudgal and Wouhaybi~\cite{mudgal2023chatgptlogs} systematically assess
ChatGPT on log processing, finding consistency and scalability
limitations once logs exceed the context window. Most closely related,
LLMLogAnalyzer~\cite{cai2025llmloganalyzer} proposes a clustering-based
log chatbot with a router, log recogniser, log parser, and search
tools, reporting substantial gains over ChatGPT and NotebookLM. These
systems share two limitations that block enterprise deployment: every
query traverses a large generative model regardless of complexity, and
external cloud APIs raise privacy and cost concerns at scale.
LogRouter adopts the modular routing philosophy
of~\cite{cai2025llmloganalyzer} and extends it with LLM-bypass paths,
Level-2 model-size selection, and fully on-premises GPU inference.

\subsection{Retrieval-Augmented Generation}
RAG~\cite{lewis2020rag} couples a neural retriever with a generative
model to improve factual grounding. Dense passage retrieval, and later
hybrid extensions, established vector similarity search as the dominant
retrieval backbone. RAG has been applied to open-domain QA, code
understanding, and enterprise knowledge bases, but its use on
operational logs remains nascent and largely offline. Our architecture
targets the distributed-ingestion, query-cost, and structural
heterogeneity gaps in the context of a production big data platform.

\subsection{Query Routing and Cost-Aware LLM Inference}
Routing queries to backends of different cost has become an active line
of work in LLM serving.
FrugalGPT~\cite{chen2023frugalgpt} cascades LLM APIs with an early-exit
confidence threshold, reaching 98\,\% cost reduction with no accuracy
loss. Adaptive-RAG~\cite{jeong2024adaptiverag} trains a lightweight
classifier to route among no-retrieval, single-step, and multi-step
retrieval, matching always-expensive baselines at lower cost. Both
share a structural assumption: every query ultimately reaches a
generative LLM. Neither considers \emph{LLM-bypass paths}---routes
that skip generation in favour of a structured database query. In log
analytics, a substantial fraction of operator questions are factual
keyword lookups or template aggregations that a time-series database
can answer exactly; bypass paths are essential. LogRouter introduces
LLM-bypass multi-path routing to the log domain, with four Level-1
paths and Level-2 model-size selection in a unified framework.

\section{System Architecture}

LogRouter is deployed within TUBITAK BILGEM's national big data
platform (hereafter the \emph{host platform}).
The system has two stages: an \emph{offline indexing pipeline} that
turns raw log streams into structured and vector indexes, and a
\emph{query-time execution layer} that routes user questions across
four paths under cost-aware dispatching. The two stages are summarised
in Figures~\ref{fig:indexing} and~\ref{fig:query}.

\subsection{Platform Context}
The host platform is a cloud-native big data infrastructure operated by
TUBITAK BILGEM, the Informatics and Information Security Research Center
of the Scientific and Technological Research Council of Turkey, for
public-sector analytics workloads. It implements a Lambda architecture
with:
\begin{itemize}
    \item \textbf{Batch processing}: PySpark for historical log
    processing and ETL pipelines.
    \item \textbf{OLAP storage}: Apache Druid, a columnar time-series
    database optimised for log retention and sub-second aggregation
    over high-cardinality event data.
    \item \textbf{Query layer}: Superset for dashboards, Trino for
    ad-hoc SQL.
\end{itemize}
LogRouter integrates into this stack as a semantic query service,
consuming the same Loki~$\rightarrow$~Spark~$\rightarrow$~Druid log
streams as the host platform's observability pipeline while adding a
\texttt{pgvector} retrieval path for semantic queries.

\noindent\textbf{Storage and orchestration.}
\begin{itemize}
    \item Druid serves as the keyword index (full-text search over raw
    lines) and the SQL backend (template-aware aggregation), with native
    time-based partitioning and rollup.
    \item PostgreSQL with the \texttt{pgvector} extension stores
    log-chunk embeddings for dense retrieval.
    \item Airflow orchestrates the indexing pipeline and periodic
    re-embedding jobs.
    \item All components run on Kubernetes with node affinity policies
    that co-locate GPU workloads (embedding, inference) and
    storage-heavy tasks (Druid historicals).
\end{itemize}

\noindent\textbf{Hardware configuration.}
All embedding and inference workloads run on a single GPU node with
an NVIDIA RTX~8000. Three models are served via Ollama on this node:
\texttt{nomic-embed-text} (768-dim) for log-chunk embedding;
\textbf{Qwen2.5-14B-Instruct} for simple semantic queries and as the
Answer Correctness judge; and \textbf{Qwen2.5-Coder-14B} for SQL
generation. The \textbf{Qwen3-32B} model handles complex semantic
synthesis and is scheduled exclusively because it consumes the full GPU
memory budget. Druid and PostgreSQL run on separate cluster nodes
connected via the platform's internal network.

\subsection{Indexing Pipeline}
Raw logs are pulled from Grafana Loki by a PySpark ingestion job and
normalised into a unified schema
\texttt{(ts, namespace, app, pod, container, level, line)}. After
normalisation the pipeline forks into two parallel branches that share
no state at indexing time:

\begin{itemize}
    \item \textbf{Structured branch.} A Drain3
    parser~\cite{he2017drain} runs as a Spark stage with two modes: a
    driver-side mode (\texttt{annotate\_df}) suitable for at most one
    million rows, and a distributed mode
    (\texttt{annotate\_df\_distributed}) that broadcasts a pre-trained
    read-only Drain3 state file and processes partitions in parallel via
    \texttt{mapPartitions}. Template IDs are content-addressable MD5
    prefixes of the normalised template string, which makes the IDs
    deterministic across executors. Severity levels are extracted from
    the line text and post-processed against a stack-trace heuristic so
    that an Exception-bearing line is always marked \texttt{ERROR}.
    The annotated rows and the extracted template catalogue are then
    ingested into Apache Druid, which serves the keyword and SQL paths
    at query time.

    \item \textbf{Semantic branch.} The same normalised line stream is
    grouped by source and split into chunks by a sliding window (default
    25 lines, 3-line overlap). Each chunk is embedded with
    \texttt{nomic-embed-text} via Ollama on the RTX~8000 and written to
    a \texttt{pgvector} table with associated metadata
    (\texttt{namespace}, \texttt{app}, \texttt{pod}, \texttt{level},
    \texttt{dataset}). The chunker operates directly on raw lines and
    does not depend on Drain output, so the two branches can be
    materialised independently.
\end{itemize}
Only the embedding stage requires GPU residency; the rest of the
pipeline is CPU-bound and scales horizontally on the Spark cluster.

\subsection{Query-Time Routing}
Query execution is governed by two lightweight routers. The Level-1
router classifies each incoming query into one of four paths:
\begin{itemize}
  \item \textbf{General}: direct model response without retrieval
  (triggered only by short greetings).
  \item \textbf{Keyword}: exact or regex-based lookup over raw log
  lines in Druid.
  \item \textbf{SQL}: template lookup followed by LLM-generated Druid
  SQL.
  \item \textbf{Semantic}: vector retrieval followed by
  evidence-grounded answer synthesis.
\end{itemize}

The Level-1 decision is made by counting matches against three regex
families---SQL signals (count, top-$k$, group-by, percentage),
keyword signals (P0--P6, described below), and event/template
signals---weighted by 0.4 per match and compared against per-category
thresholds. The default thresholds are
\texttt{sql\_threshold}=\texttt{keyword\_threshold}=0.3 and
\texttt{event\_threshold}=0.5. To prevent compound analytical questions
from accidentally triggering the keyword path on a sub-clause, an
additional gating regex (P7) requires the question to begin with a
fact-seeking word (\textit{what, where, which, when, how, is, are,
was, were, has, had, did, do, does}).

\subsubsection*{Level-1 Keyword Signal Vocabulary}
Table~\ref{tab:vocab} lists the seven structural patterns
\mbox{(P0--P6)} that constitute the keyword signal vocabulary, together
with the question-starter guard (P7). The vocabulary was iterated on
the development partition of LogHub and is intentionally small: each
pattern targets a clearly delimited class of fact-seeking question.

\begin{table}[!t]
\caption{Level-1 Keyword Signal Vocabulary.}
\label{tab:vocab}
\centering
\small
\begin{tabular}{@{}clp{4.2cm}@{}}
\toprule
\textbf{ID} & \textbf{Name} & \textbf{Example trigger} \\
\midrule
P0 & Search command           & ``Find lines containing error 503'' \\
P1 & Attribute lookup         & ``What is the IP address?'' \\
P2 & Action-verb lookup       & ``What module is being executed?'' \\
P3 & Gerund activity          & ``What is the service doing?'' \\
P4 & Past-event lookup        & ``What process crashed?'' \\
P5 & Conceptual meaning       & ``What does this error mean?'' \\
P6 & Timestamped event        & ``What happened at 03:28:22?'' \\
\midrule
P7 & Question-starter guard   & Requires \textit{what/where/\dots} at start \\
\bottomrule
\end{tabular}
\end{table}

\subsubsection*{Level-2 Model-Size Selection}
For queries that reach the Semantic path, a Level-2 router selects
between the 14B-class generator (Qwen2.5-14B-Instruct) and the 32B
generator (Qwen3-32B). Complexity is scored as the bounded sum of four
features:
\begin{equation}
c(q) = s_{\text{len}}(q) + s_{\text{agg}}(q) + s_{\text{temp}}(q) + s_{\text{ent}}(q),
\end{equation}
where each $s_i \in [0, 0.25]$ contributes at most one quarter of the
score: $s_{\text{len}}$ is proportional to query word count
(saturating at 40 words); $s_{\text{agg}}$ counts matches of
aggregation vocabulary (\textit{summarize, compare, root cause,
correlate}); $s_{\text{temp}}$ counts temporal-scope phrases
(\textit{over time, last 24h, between, after}); and $s_{\text{ent}}$
counts multi-entity connectors (\textit{and, versus, across, multiple,
each}). The threshold $c(q) \geq 0.5$ routes to Qwen3-32B; otherwise
the query is served by Qwen2.5-14B-Instruct.

\subsubsection*{Retrieval Strategies}
The default Semantic path uses \emph{hybrid retrieval}: each query is
issued in parallel as a dense vector query and a PostgreSQL
\texttt{plainto\_tsquery} full-text query (with the \texttt{simple}
dictionary so that log-specific tokens like IPs, hostnames, and error
codes are not stemmed away). When the question contains quoted
literals, those literals are also issued as a separate FTS query.
The three ranked lists are fused with Reciprocal Rank
Fusion~\cite{cormack2009rrf} ($k$=60) and the top-10 chunks are passed
to the generator. A pure-vector and a pure-keyword variant are
evaluated as ablation conditions in Section~\ref{sec:ablation}.

\subsubsection*{SQL Path}
For SQL-routed queries, the pipeline first attempts a template lookup
in Druid using a regex over the \texttt{template} and \texttt{line}
columns; if no template matches, it falls back to a \texttt{pgvector}
metadata search to retrieve representative templates. The retrieved
templates are passed as context to \mbox{Qwen2.5-Coder-14B}, which
generates a Druid SQL statement that is executed against the
\texttt{logs\_raw} datasource. The numerical result of the SQL query
is returned as the answer directly, with no further LLM rewriting,
preserving exact aggregate correctness.

\subsubsection*{Resilience}
Each backend is wrapped in a defensive try/except that downgrades the
response to a service-unavailable message rather than failing the
whole pipeline; the routing decision and per-stage latencies are
logged with a per-request trace ID. The Level-1 router enforces an
input-validation whitelist on the extracted search term so that the
keyword path cannot inject SQL via crafted operator queries.

\tikzset{
  arch/.style={font=\sffamily\footnotesize, align=center,
               every node/.style={align=center}},
  proc/.style={draw=cBorder, line width=0.55pt, rounded corners=2pt,
               fill=cProc,
               minimum width=24mm, minimum height=10mm, inner sep=2pt},
  store/.style={draw=cBorder, line width=0.55pt, cylinder,
                shape border rotate=90, aspect=0.3, fill=cStore,
                minimum width=20mm, minimum height=12mm, inner sep=1.5pt},
  source/.style={draw=cBorder, line width=0.55pt, rounded corners=2pt,
                 fill=cInput,
                 minimum width=24mm, minimum height=10mm, inner sep=2pt},
  router/.style={draw=cBorder, line width=0.6pt, diamond, aspect=2,
                 fill=cRouter,
                 minimum width=32mm, minimum height=16mm, inner sep=2pt},
  llm/.style={draw=cBorder, line width=0.55pt, rounded corners=2pt,
              fill=cLLM,
              minimum width=22mm, minimum height=10mm, inner sep=2pt},
  ans/.style={draw=cBorder, line width=1pt, rounded corners=2pt,
              fill=cAnswer,
              minimum width=26mm, minimum height=11mm, inner sep=2pt,
              font=\sffamily\footnotesize\bfseries},
  frame/.style={draw=cFrame, dashed, line width=0.45pt, rounded corners=3pt,
                inner sep=2.5mm},
  arr/.style={-{Stealth[length=2mm, width=1.5mm]}, semithick, cEdge},
  edge/.style={semithick, cEdge},
}

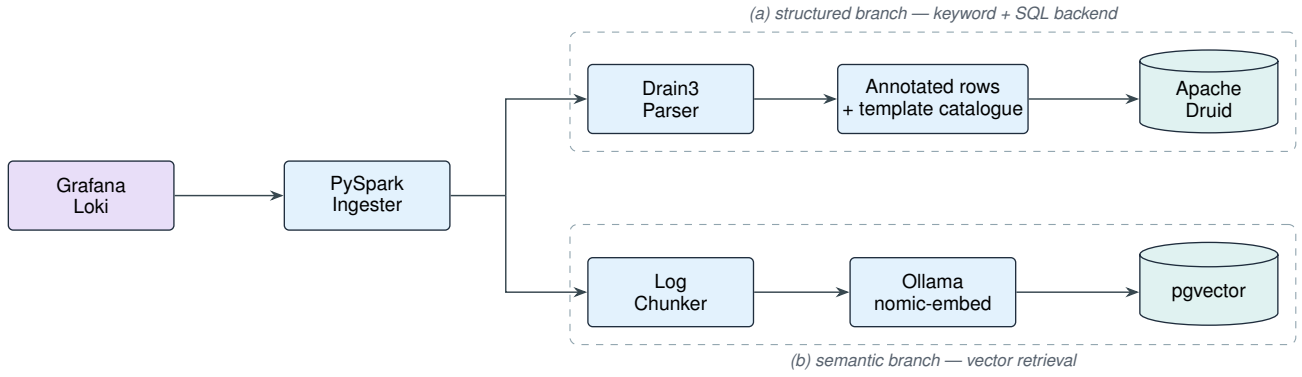
\begin{figure*}[!t]
\centering
\resizebox{0.94\textwidth}{!}{%
\begin{tikzpicture}[arch]
\node[source] (loki) at (0,0)   {Grafana\\Loki};
\node[proc]   (ing)  at (4.0,0) {PySpark\\Ingester};

\node[proc]  (drain) at (8.4, 1.4) {Drain3\\Parser};
\node[proc]  (anno)  at (12.2,1.4) {Annotated rows\\+ template catalogue};
\node[store] (druid) at (16.2,1.4) {Apache\\Druid};

\node[proc]  (chunk) at (8.4,-1.4) {Log\\Chunker};
\node[proc]  (embed) at (12.2,-1.4) {Ollama\\nomic-embed};
\node[store] (pgv)   at (16.2,-1.4) {pgvector};

\begin{scope}[on background layer]
  \node[frame, fit=(drain) (anno) (druid)] (frameS) {};
  \node[frame, fit=(chunk) (embed) (pgv)]  (frameQ) {};
\end{scope}

\node[font=\sffamily\scriptsize\itshape, text=cBorder!75,
      anchor=south] at (frameS.north) {(a) structured branch \textemdash{} keyword + SQL backend};
\node[font=\sffamily\scriptsize\itshape, text=cBorder!75,
      anchor=north] at (frameQ.south) {(b) semantic branch \textemdash{} vector retrieval};

\draw[arr] (loki) -- (ing);

\coordinate (fork) at ($(ing.east)+(0.8,0)$);
\draw[edge] (ing.east) -- (fork);
\draw[arr] (fork) |- (drain.west);
\draw[arr] (fork) |- (chunk.west);

\draw[arr] (drain) -- (anno);
\draw[arr] (anno)  -- (druid);

\draw[arr] (chunk) -- (embed);
\draw[arr] (embed) -- (pgv);
\end{tikzpicture}}
\caption{\textbf{Indexing pipeline.} Raw logs flow from Loki through
a PySpark ingester that normalises lines into a unified schema. The
normalised stream then forks into a \emph{structured branch}, where a
Drain3 parser produces annotated rows and a template catalogue ingested
into Apache Druid, and a \emph{semantic branch}, where a fixed-window
chunker feeds Ollama's \texttt{nomic-embed-text} and the resulting
vectors are stored in \texttt{pgvector}. The two branches run in
parallel and share no state at indexing time.}
\label{fig:indexing}
\end{figure*}

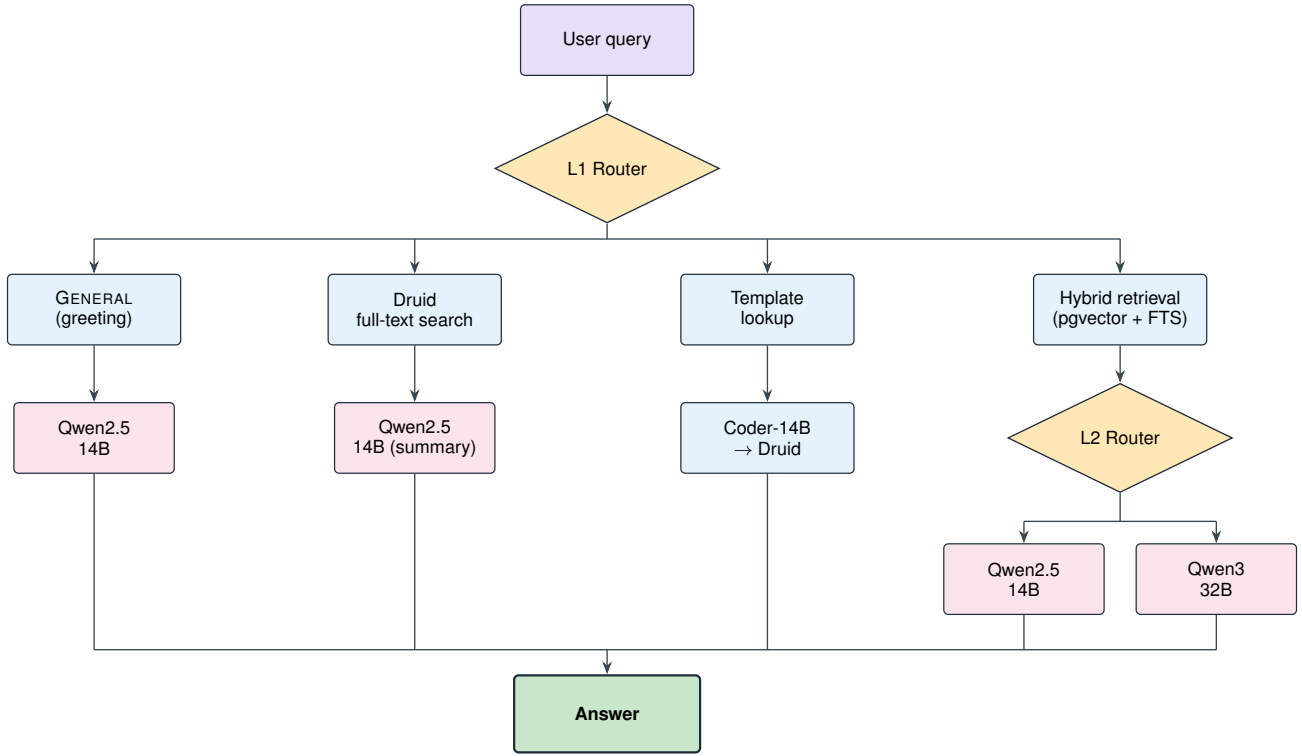
\begin{figure*}[!t]
\centering
\resizebox{0.94\textwidth}{!}{%
\begin{tikzpicture}[arch]
\tikzset{
  proc/.append style  ={minimum width=27mm, minimum height=11mm},
  source/.append style={minimum width=27mm, minimum height=11mm},
  llm/.append style   ={minimum width=25mm, minimum height=11mm},
  router/.append style={minimum width=35mm, minimum height=17mm},
  ans/.append style   ={minimum width=29mm, minimum height=12mm},
}
\node[source] (q)  at (8.0, 0.0) {User query};
\node[router] (r1) at (8.0,-2.0) {L1 Router};

\node[proc] (gen) at (0.0,-4.2)  {\textsc{General}\\(greeting)};
\node[proc] (kw)  at (5.0,-4.2)  {Druid\\full-text search};
\node[proc] (sql) at (10.5,-4.2) {Template\\lookup};
\node[proc] (sem) at (16.0,-4.2) {Hybrid retrieval\\(pgvector + FTS)};

\node[llm]    (genllm) at (0.0, -6.2) {Qwen2.5\\14B};
\node[llm]    (kwllm)  at (5.0, -6.2) {Qwen2.5\\14B (summary)};
\node[proc]   (coder)  at (10.5,-6.2) {Coder-14B\\$\rightarrow$ Druid};
\node[router] (r2)     at (16.0,-6.2) {L2 Router};

\node[llm] (m14) at (14.5,-8.4) {Qwen2.5\\14B};
\node[llm] (m32) at (17.5,-8.4) {Qwen3\\32B};

\node[ans] (ans) at (8.0,-10.5) {Answer};

\draw[arr]  (q) -- (r1);
\draw[edge] (r1.south) -- (8.0,-3.1);
\draw[edge] (0.0,-3.1) -- (16.0,-3.1);
\draw[arr]  (0.0,-3.1)  -- (gen.north);
\draw[arr]  (5.0,-3.1)  -- (kw.north);
\draw[arr]  (10.5,-3.1) -- (sql.north);
\draw[arr]  (16.0,-3.1) -- (sem.north);

\draw[arr] (gen) -- (genllm);
\draw[arr] (kw)  -- (kwllm);
\draw[arr] (sql) -- (coder);
\draw[arr] (sem) -- (r2);

\draw[edge] (r2.south) -- (16.0,-7.5);
\draw[edge] (14.5,-7.5) -- (17.5,-7.5);
\draw[arr]  (14.5,-7.5) -- (m14.north);
\draw[arr]  (17.5,-7.5) -- (m32.north);

\draw[edge] (genllm.south) -- (0.0,-9.5);
\draw[edge] (kwllm.south)  -- (5.0,-9.5);
\draw[edge] (coder.south)  -- (10.5,-9.5);
\draw[edge] (m14.south)    -- (14.5,-9.5);
\draw[edge] (m32.south)    -- (17.5,-9.5);
\draw[edge] (0.0,-9.5) -- (17.5,-9.5);
\draw[arr]  (8.0,-9.5) -- (ans.north);
\end{tikzpicture}}
\caption{\textbf{Query-time routing.} The Level-1 router dispatches
each query into one of four execution paths
(\textsc{General}, \textsc{Keyword}, \textsc{SQL}, \textsc{Semantic}).
On the semantic path the Level-2 router selects the generator size
based on a complexity score. The SQL path uses Qwen2.5-Coder-14B to
emit Druid SQL that is executed against the structured index;
its numerical result is returned without further LLM rewriting.}
\label{fig:query}
\end{figure*}

\section{Experimental Setup}
\label{sec:exp}

\subsection{Datasets and Question Sets}
We evaluate on the public \textbf{LogHub} benchmark~\cite{zhu2023loghub},
restricted to four heterogeneous system log types: Linux (19
questions), Apache (17), Windows (17), and Mac (17), for a total of
\textbf{70 questions}. Questions cover seven task families:
Summarisation, Pattern Extraction, Anomaly Detection, Root Cause
Analysis, Predictive Failure Analysis, Log Understanding/Interpretation,
and Operational Lookup. Each question is paired with a
human-written reference answer and a gold routing label
(\textsc{keyword} / \textsc{semantic} / \textsc{sql}) used to compute
per-class router metrics.

\subsection{Evaluation Modes}
We report two complementary evaluation modes that share the same
question sets:
\begin{itemize}
    \item \textbf{Full pipeline (online)}: end-to-end execution with
    live Druid and \texttt{pgvector} instances populated by the
    LogHub-indexed corpus. This mode measures the system as deployed:
    keyword answers come from Druid, SQL answers are executed against
    Druid, and semantic answers retrieve from \texttt{pgvector}.
    \item \textbf{Offline}: deterministic chunk injection without
    Druid or \texttt{pgvector}. The relevant log file is loaded from
    disk, chunked, and the top-10 chunks are passed directly to the
    generator regardless of route. This mode isolates the generator
    and router from retrieval-system variability and is used for the
    Fixed-14B and Fixed-32B baselines.
\end{itemize}

\subsection{Baselines and Ablation Conditions}
Two fixed-model baselines remove the Level-2 router by overriding the
generator size:
\begin{itemize}
  \item \textbf{Fixed-14B}: always Qwen2.5-14B-Instruct (the L2 ``small'' generator).
  \item \textbf{Fixed-32B}: always Qwen3-32B (the L2 ``large'' generator).
\end{itemize}
In addition, Section~\ref{sec:ablation} reports nine ablation
conditions: \texttt{full} (the routed system); \texttt{no-L1}
(L1 bypass, all queries routed to \textsc{semantic}); \texttt{no-L2}
(always 14B); \texttt{no-routing} (L1+L2 bypass); \texttt{semantic-only}
(no keyword backend); \texttt{keyword-only} (no semantic backend);
\texttt{hybrid} (BM25-style + dense RRF fusion);
\texttt{always-32B} (always Qwen3-32B); and \texttt{no-drain} (no
template-derived context).

\subsection{Metrics}
We adopt seven complementary metrics covering routing, lexical
overlap, semantic similarity, retrieval, generation, and latency:
\begin{itemize}
\item \textbf{Routing accuracy} with per-class precision, recall, and
$F_1$ over the gold \{keyword, semantic, sql\} labels.
\item \textbf{Cosine similarity} between sentence embeddings of the
generated and reference answers.
\item \textbf{ROUGE-1 $F_1$}~\cite{lin2004rouge}: unigram-overlap
$F_1$, sensitive to exact tokens, particularly informative for short
fact-seeking responses.
\item \textbf{BERTScore $F_1$}~\cite{zhang2020bertscore}: token-level
contextual similarity, tolerant of paraphrase and synonym variation.
\item \textbf{RAGAS Faithfulness}~\cite{es2023ragas}: fraction of
atomic claims in the generated answer that are grounded in the
retrieved context; 1.0 means every stated fact is supported.
\item \textbf{RAGAS Context Precision}~\cite{es2023ragas}: fraction of
retrieved chunks that are relevant to the question, isolating
retrieval precision from generation quality.
\item \textbf{Answer Correctness}: a local LLM judge score
(Qwen2.5-14B-Instruct as judge) that compares generated and reference
answers on a $[0,1]$ scale, combining factual overlap and semantic
similarity. The judge is sensitive to stylistic differences between
verbatim keyword-bypass responses and synthesised reference answers
(see Section~\ref{sec:ablation}).
\item \textbf{Retrieval metrics}: Hit@$k$, Recall@$k$, and MRR with
$k$=10 over the offline log-chunk candidate set.
Hit@$k$ is 1 iff any top-$k$ chunk contains the reference text.
\item \textbf{End-to-end latency}: wall-clock time from query
submission to answer delivery, including routing, retrieval, and
generation.
\end{itemize}

\subsection{Hardware and Implementation}
All experiments run on a single Kubernetes node with one NVIDIA
RTX~8000 GPU; Druid and PostgreSQL run on separate nodes. Ollama hosts
\texttt{nomic-embed-text} (embeddings, 768-d), Qwen2.5-14B-Instruct,
Qwen3-32B, and Qwen2.5-Coder-14B (generators); the 32B model occupies
the full GPU memory budget and is scheduled exclusively. The Level-1
router uses default thresholds \texttt{sql/keyword\_threshold}~$=0.3$
and \texttt{event\_threshold}~$=0.5$; the Level-2 router uses
\texttt{complexity\_threshold}~$=0.5$. Chunking uses 25-line windows
with 3-line overlap. Retrieval returns the top 10 chunks (or top
20 candidates per backend prior to RRF fusion).

\section{Results}
\label{sec:results}

\subsection{Full-Pipeline (Online) Results}
\label{sec:full_pipeline}

Table~\ref{tab:full_pipeline} reports end-to-end results across the
four LogHub datasets with the full online pipeline. The router
achieves a mean accuracy of \textbf{88.4\,\%} (62/70 questions), with
94.7\,\% on Linux. Answer quality is consistently strong across
datasets: mean BERTScore $F_1$ is 0.879 (range 0.855--0.890), mean
ROUGE-1 is 0.373, mean RAGAS Faithfulness is 0.779, and the LLM-judge
Answer Correctness averages 0.742. End-to-end latency averages
\textbf{18.6\,s}: Apache and Windows are fastest
(13.1--15.2\,s) because a larger share of their questions resolve on
the keyword or SQL paths that bypass the 32B generator; Mac is
slowest (23.6\,s) because its complex semantic questions disproportionately
trigger the Level-2 router into the 32B path.

The relationship between routing accuracy and answer quality is direct.
Linux---with the highest routing accuracy---also yields the highest
Cosine, ROUGE-1, BERTScore, and Answer Correctness. The few percent
loss on Apache, Windows, and Mac stems almost entirely from
keyword-to-semantic over-routing: questions phrased like attribute
lookups (P1) sometimes belong to the semantic path, and forcing them
onto the keyword path produces a single matched line instead of the
synthesised paragraph that the reference answer expects.

\begin{table*}[!t]
\caption{Full-Pipeline (Online) Results Across LogHub Datasets (70 Total Questions).}
\label{tab:full_pipeline}
\centering
\small
\setlength{\tabcolsep}{6pt}
\begin{tabular}{@{}lrcccccccc@{}}
\toprule
\textbf{Dataset} & \textbf{N} & \textbf{Router Acc} & \textbf{Cosine} & \textbf{ROUGE-1} & \textbf{BERTScore} & \textbf{RAGAS Faith.} & \textbf{Ctx Prec.} & \textbf{Correct.} & \textbf{Lat. (s)} \\
\midrule
Linux   & 19 & 0.947          & \textbf{0.853} & \textbf{0.434} & \textbf{0.890} & 0.741          & 0.900          & \textbf{0.828} & 22.5 \\
Apache  & 17 & 0.824          & 0.832          & 0.393          & 0.883          & 0.766          & \textbf{1.000} & 0.766          & \textbf{13.1} \\
Windows & 17 & 0.882          & 0.798          & 0.388          & 0.886          & 0.760          & 0.747          & 0.708          & 15.2 \\
Mac     & 17 & 0.882          & 0.753          & 0.277          & 0.855          & \textbf{0.850} & 0.700          & 0.668          & 23.6 \\
\midrule
\textbf{Mean} & \textbf{70} & \textbf{0.884} & \textbf{0.809} & \textbf{0.373} & \textbf{0.879} & \textbf{0.779} & \textbf{0.837} & \textbf{0.742} & \textbf{18.6} \\
\bottomrule
\multicolumn{10}{@{}l@{}}{\footnotesize Bold marks the best per-column value. Latency is wall-clock end-to-end.}
\end{tabular}
\end{table*}

\subsection{Routing Accuracy and Confusion Analysis}
\label{sec:routing}

Table~\ref{tab:linux_routing} reports per-class metrics for the Linux
benchmark, our highest-routing dataset. The keyword signal vocabulary
(P0--P7) classifies 18 of 19 questions correctly, with keyword $F_1$
of 0.909, semantic $F_1$ of 0.957, and SQL $F_1$ of 1.0. The single
remaining error is \texttt{linux\_q8} (``What does each column in this
log mean?''), which triggers the conceptual-meaning pattern P5 even
though it is a \emph{schema-level} question with no matching log line
to retrieve. P5 is therefore a residual false-positive source we revisit
in the Limitations section.

\begin{table}[!t]
\caption{Per-Class Routing Results on Linux (full pipeline).}
\label{tab:linux_routing}
\centering
\small
\begin{tabular}{@{}lcccc@{}}
\toprule
\textbf{Class} & \textbf{Supp.} & \textbf{Prec.} & \textbf{Rec.} & \textbf{F1} \\
\midrule
Keyword  & \phantom{0}5 & 0.833 & 1.000 & 0.909 \\
Semantic & 12           & 1.000 & 0.917 & 0.957 \\
SQL      & \phantom{0}2 & 1.000 & 1.000 & 1.000 \\
\bottomrule
\end{tabular}
\end{table}

\begin{figure}[!t]
  \centering
  \includegraphics[width=\linewidth]{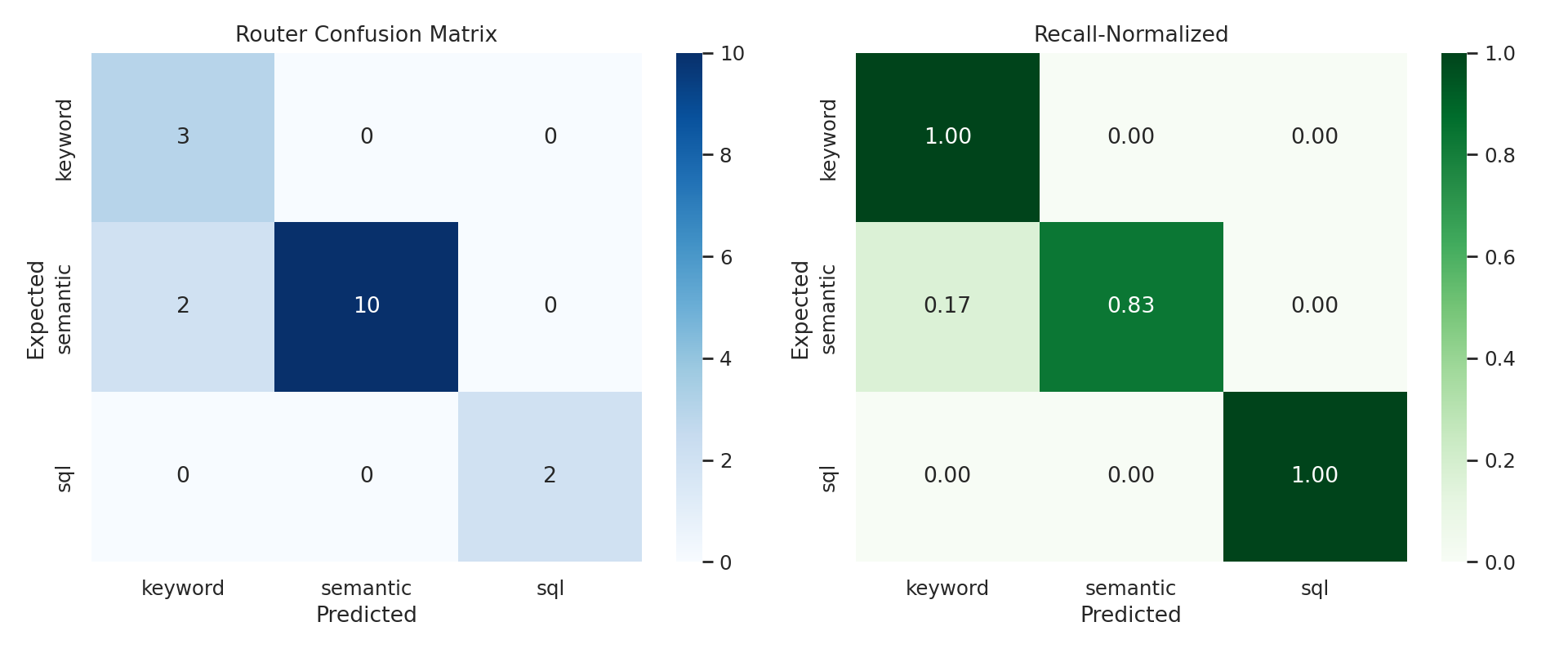}
  \caption{Router confusion matrix (left) and recall-normalised view
  (right) aggregated over the four LogHub datasets. SQL is classified
  perfectly; the residual error is semantic-to-keyword over-routing.}
  \label{fig:router_conf}
\end{figure}

Across all four datasets, the dominant remaining error pattern is
\emph{semantic-to-keyword} over-routing, in which two
``what is the behaviour of X?'' style questions are routed to the
keyword path because P1/P2 fire on the literal entity term while the
question semantically requires multi-line synthesis. The aggregate
confusion matrix over the four datasets (Figure~\ref{fig:router_conf})
shows perfect SQL classification and a small residual flow of
keyword-style predictions on semantic questions; no semantic queries
are misrouted to the SQL path and no SQL queries are misrouted at all.

\subsection{Offline Baseline Comparison}
\label{sec:baselines}

Table~\ref{tab:baselines} compares the routed system against Fixed-14B
and Fixed-32B baselines in the offline configuration (raw chunk
injection, no Druid/\texttt{pgvector}). Three findings stand out.

\textbf{Latency.} The routed system reduces mean end-to-end latency by
\textbf{55\,\%} versus Fixed-32B (46.3\,s vs.\ 102.1\,s) and is on par
with Fixed-14B (46.3\,s). The Mac dataset shows the largest gap:
73.8\,s (routed) vs.\ 136.1\,s (Fixed-32B), because Mac contains the
longest semantic questions and Qwen3-32B amortises the slowest on
those. The Level-2 router invokes the 32B model only when query
complexity warrants it, which is the primary cost-reduction mechanism.

\textbf{Faithfulness.} The routed system reaches a mean RAGAS
Faithfulness of \textbf{0.858}---higher than Fixed-14B (0.799) and
lower than Fixed-32B (0.918). On a per-dataset basis the routed system
exceeds Fixed-14B on every dataset (notably Mac: 0.744 vs.\ 0.625) and
recovers most of the Fixed-32B faithfulness gap at less than half the
cost. Fixed-32B is the strongest in absolute faithfulness because the
larger model hallucinates less under multi-evidence synthesis, but it
does so at a $2.2\times$ latency penalty.

\textbf{Answer Correctness.} Mean LLM-judge Correctness is
\textbf{0.726} (routed) versus 0.716 (Fixed-14B) and 0.784
(Fixed-32B). The 5.8-point gap to Fixed-32B is concentrated on Apache
and Mac, where the 32B model's elaborated paragraphs better match the
length of the reference answers; on Linux and Windows the routed system
is within 3 points of Fixed-32B at less than half the latency. The
Correctness metric reflects stylistic closeness to the reference; a
brief, verbatim keyword-bypass answer is penalised even when it is
factually exact.

\textbf{BERTScore.} The three systems are nearly indistinguishable in
BERTScore (routed: 0.856, Fixed-14B: 0.857, Fixed-32B: 0.846); the 32B
model's verbosity slightly hurts BERTScore by extending answers beyond
the reference length.

\begin{table*}[!t]
\caption{Offline Baseline Comparison Across Four LogHub Datasets (17--19 questions each, 70 total).}
\label{tab:baselines}
\centering
\small
\setlength{\tabcolsep}{8pt}
\begin{tabular}{@{}llcccc@{}}
\toprule
\textbf{Dataset} & \textbf{System} & \textbf{BERTScore F1\,$\uparrow$} & \textbf{RAGAS Faith.\,$\uparrow$} & \textbf{Answer Correct.\,$\uparrow$} & \textbf{Latency (s)\,$\downarrow$} \\
\midrule
\multirow{3}{*}{Linux}
  & Routed (ours) & 0.870          & 0.930          & 0.763          & 43.3 \\
  & Fixed-14B     & \textbf{0.877} & 0.908          & 0.745          & \textbf{42.2} \\
  & Fixed-32B     & 0.852          & \textbf{0.968} & \textbf{0.792} & 99.2 \\
\midrule
\multirow{3}{*}{Apache}
  & Routed (ours) & \textbf{0.870} & 0.824          & 0.712          & \textbf{30.3} \\
  & Fixed-14B     & 0.866          & 0.794          & 0.703          & 31.0 \\
  & Fixed-32B     & 0.861          & \textbf{0.985} & \textbf{0.797} & 99.4 \\
\midrule
\multirow{3}{*}{Windows}
  & Routed (ours) & 0.865          & \textbf{0.934} & 0.753          & 37.7 \\
  & Fixed-14B     & \textbf{0.866} & 0.871          & 0.747          & \textbf{37.5} \\
  & Fixed-32B     & 0.840          & 0.871          & \textbf{0.806} & 73.8 \\
\midrule
\multirow{3}{*}{Mac}
  & Routed (ours) & 0.818          & 0.744          & 0.677          & \textbf{73.8} \\
  & Fixed-14B     & 0.818          & 0.625          & 0.671          & 74.6 \\
  & Fixed-32B     & \textbf{0.831} & \textbf{0.848} & \textbf{0.741} & 136.1 \\
\midrule
\multirow{3}{*}{\textbf{Mean}}
  & Routed (ours) & 0.856          & 0.858          & 0.726          & 46.3 \\
  & Fixed-14B     & \textbf{0.857} & 0.799          & 0.716          & \textbf{46.3} \\
  & Fixed-32B     & 0.846          & \textbf{0.918} & \textbf{0.784} & 102.1 \\
\bottomrule
\multicolumn{6}{@{}l@{}}{\footnotesize Bold: best value per group. Fixed-14B refers to Qwen2.5-14B-Instruct (the L2 ``small'' generator).}
\end{tabular}
\end{table*}

\subsection{Ablation Study}
\label{sec:ablation}

\begin{figure}[!t]
  \centering
  \includegraphics[width=\linewidth]{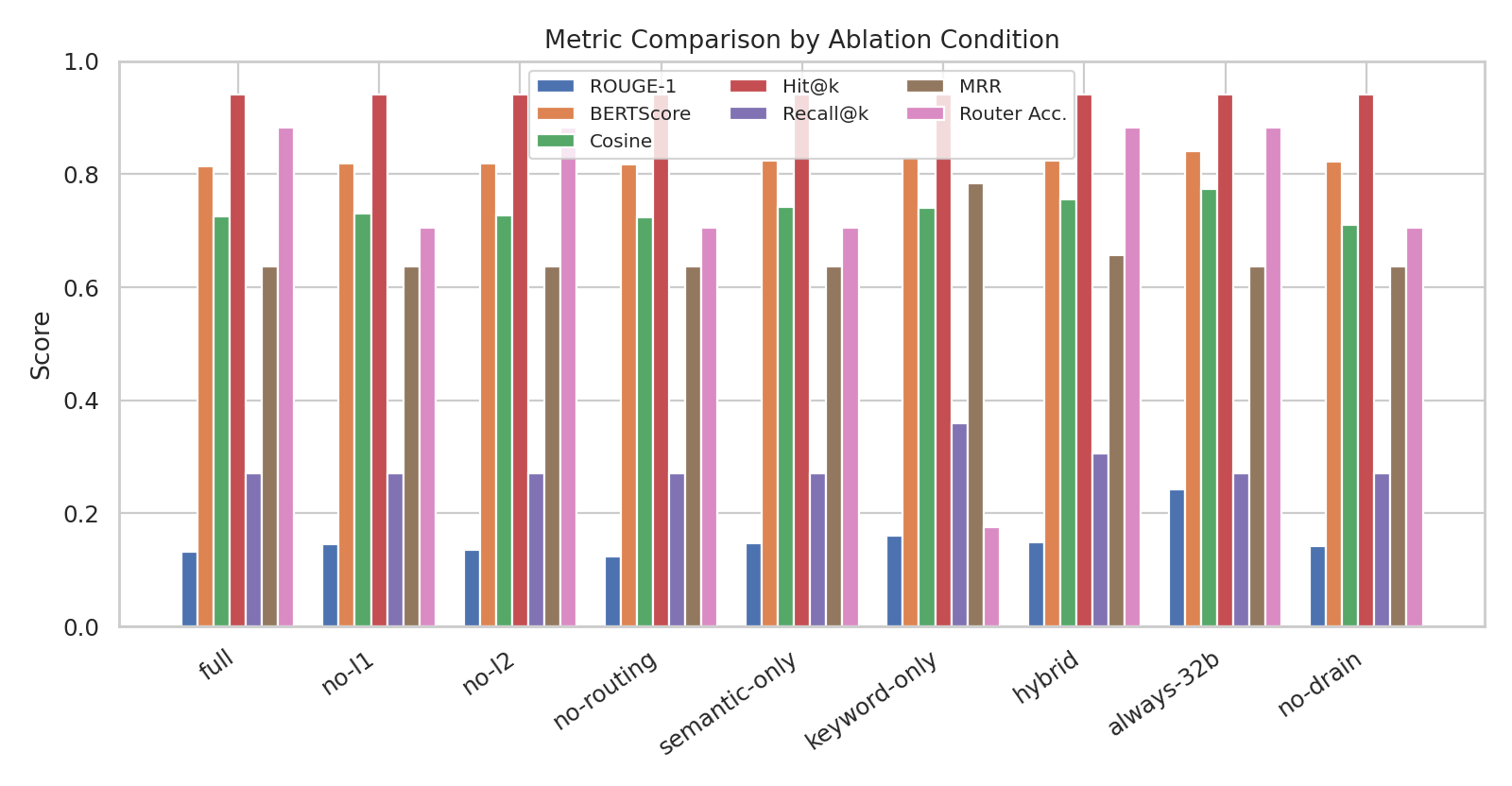}
  \caption{Metric comparison across ablation conditions, averaged over
  the four LogHub datasets. Removing L1, routing, or Drain collapses
  routing accuracy without recovering quality; \texttt{always-32B}
  matches the full system's routing but costs more than $2.5\times$
  the latency.}
  \label{fig:ablation_metrics}
\end{figure}

Table~\ref{tab:ablation} reports the nine-condition ablation averaged
over the four LogHub datasets, and Figure~\ref{fig:ablation_metrics}
visualises the same conditions side by side. Several conclusions
emerge.

\textbf{(1) Drain is the single most important component for routing
accuracy.} Removing template-derived context (\texttt{no-drain}) drops
routing accuracy by 19.7 percentage points (0.884$\rightarrow$0.687),
matching the drop produced by removing the L1 router entirely. Without
templates the L1 keyword vocabulary lacks the normalised tokens it
relies on, and the system silently falls back to semantic for every
query. Generation quality is essentially unchanged
(BERTScore 0.851 vs.\ 0.849), so the cost of removing Drain is paid
entirely in misrouting and a 54\,\% latency penalty
(29.4\,s vs.\ 19.1\,s)---most queries then traverse the more expensive
semantic path unnecessarily.

\textbf{(2) The Level-2 router is a near-free quality knob.}
\texttt{no-L2} retains the same routing accuracy as the full system
(0.884) but routes every semantic query to the 14B generator. BERTScore
and Cosine are virtually identical to the full system, while latency
drops marginally (17.4\,s vs.\ 19.1\,s). The full system is therefore
spending a small latency budget at L2 to gain its larger faithfulness
margin (see Section~\ref{sec:baselines}, where the routed system
exceeds Fixed-14B on faithfulness on every dataset). Conversely,
\texttt{always-32B} achieves the same routing accuracy and slightly
higher Cosine (0.756 vs.\ 0.739) but raises mean latency by
\textbf{160\,\%} (49.6\,s vs.\ 19.1\,s). The L2 router occupies a
favourable efficient frontier between these extremes.

\textbf{(3) Routing accuracy and LLM-judge correctness can diverge.}
\texttt{no-L1}, \texttt{no-routing}, and \texttt{semantic-only} all
collapse routing accuracy to 0.687 yet leave ROUGE-1, BERTScore, and
Cosine essentially unchanged. The judge rewards stylistic closeness:
when a keyword-style question is misrouted to the semantic path, the
generator produces a synthesised paragraph that better matches the
reference's prose, even though the answer is factually less precise
than the bypass response. This is a structural artefact of LLM-judge
scoring rather than a real quality regression---retrieval-grounded
metrics (RAGAS Faithfulness, exact retrieval Hit@$k$) capture the
underlying precision that the judge misses.

\textbf{(4) Hybrid retrieval extends recall at modest cost.}
\texttt{hybrid} (RRF fusion of vector and BM25-style scoring) achieves
the highest mean Recall@$k$ (0.269 vs.\ 0.154 for the full system) and
the second-highest Cosine (0.759), confirming that combining dense and
lexical signals enlarges the recall envelope on fact-seeking
sub-questions. Latency is 33.8\,s---higher than \texttt{full} because
the system runs two retrieval passes, but lower than \texttt{always-32B}.
\texttt{keyword-only} is the absolute Recall winner (0.361), but its
routing accuracy collapses to 0.198 because every query is forced
through the keyword path---a useful diagnostic but not a deployable
configuration.

\begin{figure}[!t]
  \centering
  \includegraphics[width=\linewidth]{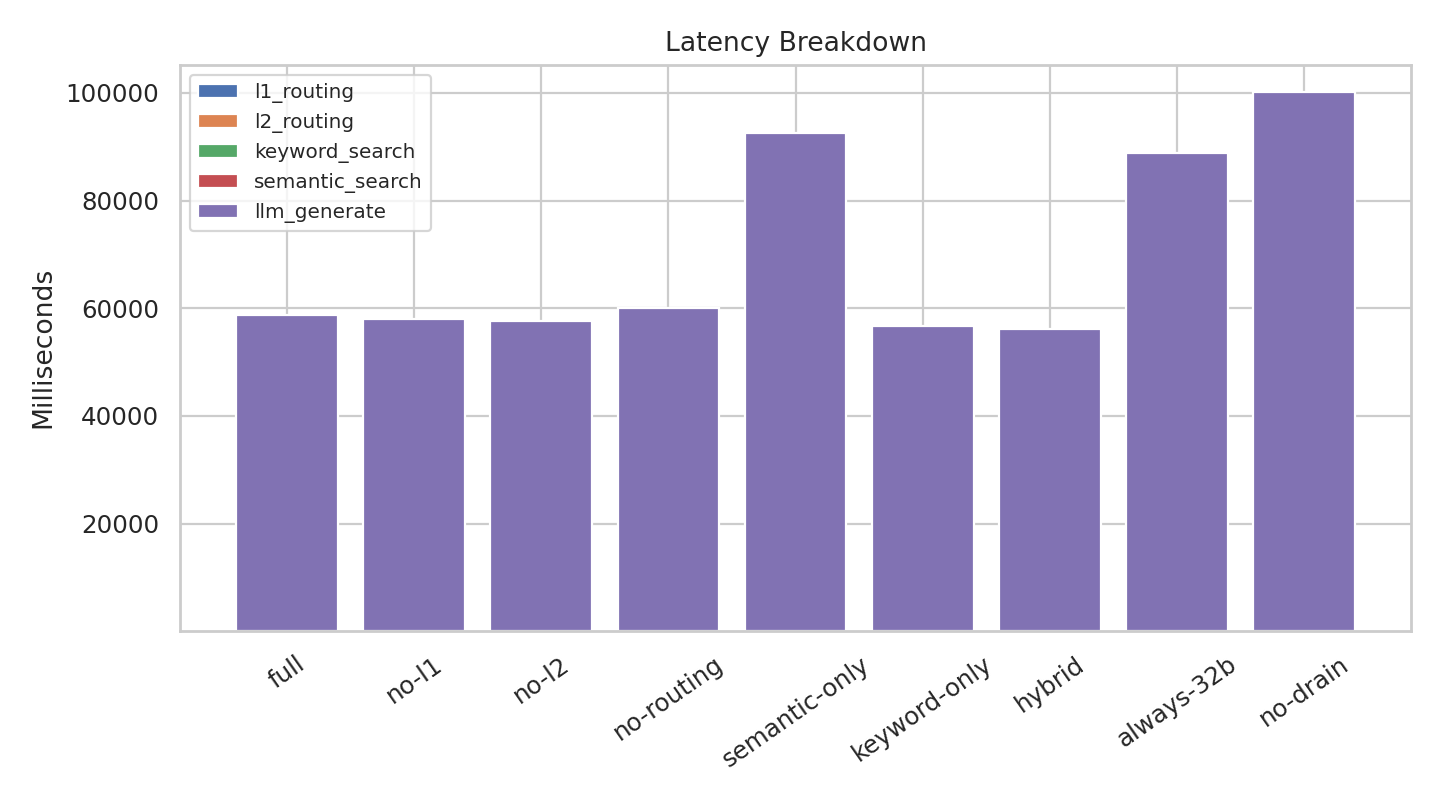}
  \caption{Per-stage latency breakdown by ablation condition. The
  \texttt{llm\_generate} stage dominates; routing and retrieval are
  collectively sub-second. The full system spends an order of
  magnitude less time in generation than \texttt{always-32B}.}
  \label{fig:latency_breakdown}
\end{figure}

\textbf{(5) Routing is what saves latency; retrieval barely registers.}
Figure~\ref{fig:latency_breakdown} stacks the per-stage latency
breakdown across all conditions. The
\texttt{llm\_generate} stage dominates every condition; L1 routing, L2
routing, keyword search, and semantic search are collectively
sub-second on a single GPU node. The latency gap between
\texttt{always-32B} (49.6\,s) and \texttt{full} (19.1\,s) is therefore
attributable entirely to model selection, not retrieval engineering.

\begin{table*}[!t]
\caption{Ablation Study: Mean Metrics Across Four LogHub Datasets.}
\label{tab:ablation}
\centering
\small
\setlength{\tabcolsep}{6pt}
\begin{tabular}{@{}lcccccc@{}}
\toprule
\textbf{Condition} & \textbf{Router Acc\,$\uparrow$} & \textbf{BERTScore\,$\uparrow$} & \textbf{Cosine\,$\uparrow$} & \textbf{ROUGE-1\,$\uparrow$} & \textbf{Recall@$k$\,$\uparrow$} & \textbf{Lat. (s)\,$\downarrow$} \\
\midrule
Full (ours)     & \textbf{0.884} & 0.849          & 0.739          & 0.233          & 0.154          & \textbf{19.1} \\
\midrule
no-L1           & 0.687          & 0.850          & 0.740          & 0.242          & 0.154          & 17.6 \\
no-L2           & \textbf{0.884} & 0.851          & 0.740          & 0.238          & 0.154          & 17.4 \\
no-routing      & 0.687          & 0.850          & 0.738          & 0.232          & 0.154          & 18.1 \\
semantic-only   & 0.687          & 0.851          & 0.744          & 0.243          & 0.154          & 26.1 \\
keyword-only    & 0.198          & \textbf{0.857} & \textbf{0.767} & \textbf{0.262} & \textbf{0.361} & 37.3 \\
hybrid          & \textbf{0.884} & 0.853          & 0.759          & 0.250          & 0.269          & 33.8 \\
always-32B      & \textbf{0.884} & 0.847          & 0.756          & 0.249          & 0.154          & 49.6 \\
no-drain        & 0.687          & 0.851          & 0.735          & 0.236          & 0.154          & 29.4 \\
\bottomrule
\multicolumn{7}{@{}l@{}}{\footnotesize Bold: best value per column. Recall@$k$ computed over the offline log-chunk candidate set.}
\end{tabular}
\end{table*}

\subsection{Per-Route Answer Quality}
Within the Linux full-pipeline run, answer quality differs sharply by
route. Correctly-routed semantic questions average Cosine 0.778 and
ROUGE-1 $F_1$ 0.175. The two keyword-misrouted semantic questions
average Cosine 0.576 and ROUGE-1 $F_1$ 0.113: the keyword bypass
returns a single matched log line, whereas the reference answer is a
paragraph---surface-form metrics penalise this mismatch, even though
the matched line is itself factually relevant. SQL-path answers
return in under 4\,s, semantic-path answers average 26.6\,s on the
14B generator (longer when the L2 router invokes the 32B model), and
keyword-bypass answers return in under 1\,s. Bypass routing is
therefore both the latency floor of the system and the path with the
highest format mismatch to synthesised references.

\section{Discussion}

The results yield three architectural lessons for production log
QA systems.

\textbf{Routing is a latency lever, not a quality penalty.} The
offline comparison in Table~\ref{tab:baselines} shows that the routed
system recovers most of the Fixed-32B faithfulness gain
(0.858 vs.\ 0.918) while halving its latency
(46.3\,s vs.\ 102.1\,s). Compared to Fixed-14B, the routed system is
faster on Apache, Windows, and Mac and strictly more faithful on every
dataset. The cost-aware router is therefore Pareto-improving against
the smaller fixed baseline and a strong latency-quality trade-off
against the larger one.

\textbf{Drain templates buy routing, not generation.} The
\texttt{no-drain} ablation drops routing accuracy by 19.7 points but
leaves BERTScore unchanged. This isolates the role of template
extraction in the system: it normalises log vocabulary so the L1
router can match keyword patterns reliably, but it does not directly
contribute to generation quality once a question reaches the correct
path. Practitioners can deprioritise Drain accuracy tuning for
generation purposes but should not skip Drain altogether.

\textbf{LLM-judge correctness rewards style.} The ablation conditions
that destroy routing accuracy (\texttt{no-L1}, \texttt{no-routing},
\texttt{semantic-only}) maintain or even slightly improve LLM-judge
correctness. This is a quiet but consequential measurement artefact:
when keyword-style questions are misrouted to the semantic path the
generator returns longer, more reference-shaped paragraphs, even when
the underlying fact is less precisely identified. RAGAS Faithfulness
and routing accuracy are therefore the metrics that align with
operational answer quality, and we use them as the primary criteria
for production tuning. Future work should treat LLM-judge scores with
explicit style-correction or pair them with factual-precision
oracles.

\section{Limitations and Future Work}

\begin{itemize}
\item \textbf{Keyword recall ceiling.} The router correctly handles
patterns P0--P6 but misses naturally phrased questions outside the
vocabulary: \textit{why}-phrased questions describing directly
observable log events and \textit{what}-phrased questions with
compositional structures (e.g., conjoined sub-questions). A learned
intent classifier trained on the misrouted set would likely close this
gap without sacrificing the high precision of the pattern-based
approach.

\item \textbf{P5 false positive on schema questions.}
The P5 pattern (\texttt{what does X mean}) misclassifies schema-level
questions (e.g., \texttt{linux\_q8}: ``What does each column in this
log mean?'') as keyword lookups. A simple fix is to require P5 to be
shorter than $N$ tokens or to include a vocabulary-based exclusion for
schema vocabulary (\textit{column, field, row, schema, format}).

\item \textbf{SQL routing precision.} SQL routing precision is 66.7\,\%
in the full pipeline: one in three SQL-path predictions arrives from a
non-SQL question, causing an unnecessary SQL-generation attempt. The
cause is broad temporal-phrase matching in the SQL signal; tightening
the pattern to require explicit aggregation vocabulary alongside a
temporal scope term would improve precision without losing the
high recall observed in our SQL test set.

\item \textbf{Hybrid context precision.} Hybrid retrieval improves
mean Recall@$k$ (0.269 vs.\ 0.154 for the full system) but
introduces additional chunks that reduce RAGAS Context Precision. A
cross-encoder re-ranker between retrieval and generation would filter
noise while retaining the recall gains.

\item \textbf{Hardware constraints and model capacity.} Self-hosting
on a single RTX~8000 caps the largest generator at Qwen3-32B and
forces exclusive scheduling. Future work will evaluate the LogRouter
framework with larger on-premises models (e.g.\ Llama-3.1-70B) to
test whether the Level-2 router still occupies the efficient frontier
when the ``large'' path becomes substantially more capable.

\item \textbf{Ingestion scalability and ecosystem integration.}
The Grafana Loki ingestion path is sufficient for the current host
platform workload but limits horizontal throughput. A planned
transition to Apache Kafka + Spark Streaming will enable
backpressure-controlled, real-time ingestion at scale and unify the
streaming path with the rest of the host platform's infrastructure.
\end{itemize}

\section{Conclusion}

We presented LogRouter, an end-to-end, self-hosted log
question-answering system built from production-grade observability
components: Grafana Loki for ingestion, PySpark + Drain3 for template
extraction, \texttt{pgvector} for semantic recall, Apache Druid for
keyword and SQL execution, and local Qwen-family models for generation.
A two-level cost-aware router dispatches each query across four
execution paths and selects a 14B-class or 32B-class generator only
when query complexity warrants the larger model.

Across four LogHub datasets the router reaches 88.4\,\% mean accuracy
(94.7\,\% on Linux); the full pipeline attains a mean ROUGE-1 of
0.373, BERTScore of 0.879, and end-to-end latency of 18.6\,s. The
routed system recovers most of the faithfulness of an always-32B
configuration while halving its latency, and it strictly dominates the
always-14B baseline on faithfulness across every dataset. The nine-condition
ablation isolates Drain template extraction as the dominant
component for routing accuracy, the Level-2 router as a near-free
quality knob, and hybrid retrieval as the cheapest recall lever
available.

Beyond the specific deployment, the results support a broader
architectural claim: in cost-constrained, self-hosted log analytics,
cost-aware routing is not a quality compromise---it is the mechanism
that makes high-quality on-premises log QA economically viable.

\section*{Disclosure on AI Use}
In line with IEEE policy on AI-generated content, we disclose that
\textbf{Claude} (Anthropic) was used as a writing assistant in the
preparation of this manuscript. Its involvement was limited to
refining wording and restructuring passages for clarity in the
narrative sections (Abstract, Introduction, Related Work, Discussion,
and Limitations); no AI system contributed to the system design,
experimental setup, evaluation, or the production of any results,
figures, or tables. All technical claims, statistical analyses, and
quantitative findings reported in this paper were produced and
verified by the authors, who are solely responsible for the
correctness and integrity of the content.

\section*{Open Science}
To support replication, the implementation, evaluation harness, the
LogHub-derived question set with gold routing labels, and the
hyper-parameter configurations used to produce the results in this
paper will be released under an open-source licence in a public
repository upon acceptance.

\bibliographystyle{IEEEtran}
\bibliography{references}

\end{document}